\documentclass[
    sigconf,
    balance=false,
    pbalance=true
]{acmart}

\usepackage{hyperref}
\usepackage{url}

\AtBeginDocument{%
  }

\copyrightyear{2026}
\acmYear{2026}
\setcopyright{cc}
\setcctype{by}
\acmConference[MM '26]{Proceedings of the 34th ACM International Conference on Multimedia}{November 10--14, 2026}{Rio de Janeiro, Brazil}
\acmBooktitle{Proceedings of the 34th ACM International Conference on Multimedia (MM '26), November 10--14, 2026, Rio de Janeiro, Brazil}
\acmDOI{10.1145/3767308.3838656}
\acmISBN{979-8-4007-2213-4/2026/11}

\usepackage[table]{xcolor}
\definecolor{deepgreen}{RGB}{0,128,0}
\definecolor{deepred}{RGB}{200,0,0}
\usepackage[utf8]{inputenc} % allow utf-8 input
\usepackage[T1]{fontenc}    % use 8-bit T1 fonts
\usepackage{hyperref}       % hyperlinks
\usepackage{url}            % simple URL typesetting
\usepackage{booktabs}       % professional-quality tables
\usepackage{amsfonts}       % blackboard math symbols

\usepackage{nicefrac}       % compact symbols for 1/2, etc.
\usepackage{microtype}      % microtypography
\usepackage{xcolor}         % colors
\usepackage{graphicx}
\usepackage{subcaption}
\usepackage{adjustbox}
\usepackage{graphicx}
\usepackage{xspace}
\usepackage{wrapfig}
\usepackage{float}
\usepackage{amsmath}
\usepackage{multirow}
\usepackage{multirow}
\usepackage{tcolorbox}
\usepackage{algorithmicx}
\usepackage{algorithm}% http://ctan.org/pkg/algorithms
\usepackage{algpseudocode}% http://ctan.org/pkg/algorithmicx
\newcommand{\KG}{{\textsc{MedMKG}\xspace}}

\usepackage{enumitem}
\usepackage[T1]{fontenc}
\usepackage[normalem]{ulem}

\usepackage{lineno}
\title{\KG: Benchmarking Medical Knowledge Exploitation with Multimodal Knowledge Graph}

\author{Xiaochen Wang}
  \affiliation{%
    \institution{The Pennsylvania State University}
    \city{University Park}
    \state{PA}
    \country{USA}
  }%
\email{xcwang@psu.edu}

\author{Yuan Zhong}
  \affiliation{%
    \institution{The Pennsylvania State University}
    \city{University Park}
    \state{PA}
    \country{USA}
  }%
\email{yfz5556@psu.edu}

\author{Lingwei Zhang}
  \affiliation{%
    \institution{The Pennsylvania State University}
    \city{University Park}
    \state{PA}
    \country{USA}
  }%
\email{lingwei@psu.edu}

\author{Lisong Dai}
  \affiliation{%
    \institution{Renmin Hospital of Wuhan University}
    \city{Wuhan}
    \state{Hubei}
    \country{China}
  }%
\email{lisong-dai@whu.edu.cn}

\author{Ting Wang}
  \affiliation{%
    \institution{Stony Brook University}
    \city{Stony Brook}
    \state{NY}
    \country{USA}
  }%
\email{twang@cs.stonybrook.edu}

\author{Fenglong Ma}
\authornote{Corresponding author.}
  \affiliation{%
    \institution{The Pennsylvania State University}
    \city{University Park}
    \state{PA}
    \country{USA}
  }%
\email{fenglong@psu.edu}

\renewcommand{\shortauthors}{Xiaochen Wang et al.}
\begin{document}

\begin{abstract}

Deep learning models in medicine increasingly depend on structured clinical knowledge for complex reasoning and diagnosis. While prior work has mainly leveraged unimodal resources such as the Unified Medical Language System (UMLS), multimodal knowledge graphs remain underexplored due to the lack of datasets linking medical images with structured clinical concepts.
To fill this gap, we introduce {\KG}, a medical multimodal knowledge graph that integrates radiological chest X-ray images from MIMIC-CXR with an ontological knowledge graph from UMLS via a multi-stage construction framework. {\KG} employs a hybrid pipeline combining rule-based extraction and large language model-based disambiguation to generate reliable concepts and relations. To improve graph quality, we propose Neighbor-aware Filtering (NaF), an efficient method for reducing redundancy during multimodal knowledge graph construction. 
We establish {\KG} as a benchmark by evaluating it on 5 medical tasks across 2 experimental settings, using 24 baseline methods and 4 vision–language backbones across 6 datasets. Results demonstrate consistent performance gains across tasks, highlighting {\KG}'s value for benchmarking and advancing multimodal knowledge integration in radiology.\footnote{Materials and appendix content are available at \url{https://github.com/XiaochenWang-PSU/MedMKG/}}

\end{abstract}




\maketitle

\section{Introduction}
\label{sec:intro}
Deep learning has demonstrated remarkable success in the medical domain, enabling tasks such as health risk prediction, disease diagnosis, and mortality forecasting~\cite{esteva2019guide, wang2024recent, miotto2018deep}. However, medical data often suffer from noise and missing values, limiting the effectiveness of feature representation learning. To address these challenges, researchers have increasingly integrated \textit{unimodal} medical knowledge graphs into deep learning frameworks. These graphs offer structured and explicit representations of domain knowledge by encoding well-defined medical concepts and their relationships~\cite{wang2025developing, qu2022review, li2020real, wu2023medical}. Incorporating such structured knowledge has led to notable improvements in different tasks, such as health risk prediction~\cite{ye2021medpath, choi2016retain, ma2020adacare}, adverse drug reaction prediction~\cite{wang2021adverse, zhang2021prediction, bean2017knowledge}, medical coding~\cite{luo2024corelation, shi2017towards}, and medical information retrieval~\cite{wang2023neurobridge, wang2023enabling}.

\begin{figure*}
    \centering
    \includegraphics[width=0.92\textwidth]{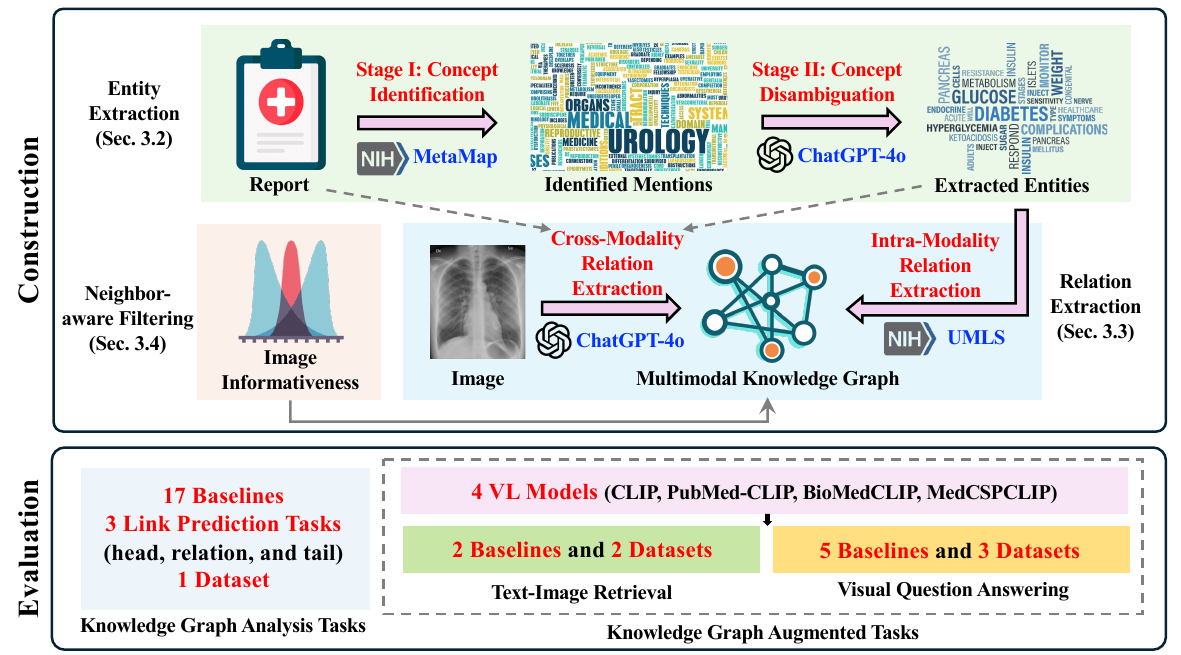}
        \vspace{-0.15in}
    \caption{The overview of {\KG} construction pipeline and evaluation methods.}
    \vspace{-0.2in}
    \label{fig:intro}
    % \vspace{-0.2in}
\end{figure*}

Nevertheless, many important clinical tasks require \textbf{multimodal} data as model inputs, such as medical visual question answering (VQA)~\cite{lin2023medical} and text-image retrieval~\cite{kitanovski2017multimodal}. Relying solely on unimodal medical knowledge graphs in these contexts often fails to yield significant performance gains, due to the absence of explicit relationships between visual data and medical concepts. This limitation has hindered the ability of current multimodal deep learning models to fully capitalize on domain knowledge in knowledge-intensive tasks.
Addressing this gap necessitates the development of a comprehensive multimodal medical knowledge graph. However, building such a resource introduces the following critical challenges:
\begin{itemize}[leftmargin=*]
\item \textbf{C1: Quality Concern}. A multimodal medical knowledge graph must be of high quality and practical utility. This includes the accurate identification and representation of diverse intra- and inter-modal relationships, which requires a carefully designed and systematically implemented construction process.

\item \textbf{C2: Utility Concern}. Beyond quality, it is essential to evaluate whether the graph can effectively enhance model performance on downstream tasks. The graph must encode clinically meaningful multimodal knowledge that directly supports a wide range of knowledge-intensive applications.
\end{itemize}

To bridge this research gap and address the identified challenges, we introduce {\KG}, a \textbf{Rad}iological \textbf{M}ultimodal \textbf{K}nowledge \textbf{G}raph that unifies visual and textual medical information. To tackle \textbf{C1} (\textit{Quality Concern}), as shown in Figure~\ref{fig:intro}, we develop a multi-stage construction pipeline that ensures high-fidelity cross-modal integration by combining the rich visual and textual information in MIMIC-CXR~\cite{johnson2019mimic} with the structured clinical knowledge in the Unified Medical Language System (UMLS)~\cite{bodenreider2004unified}. Our method leverages the domain accuracy of rule-based tools together with the contextual reasoning capabilities of large language models (LLMs), enabling precise extraction of clinical concepts and their relationships. To further ensure conciseness and informativeness, we propose a simple yet effective Neighbor-aware Filtering algorithm (NaF) to enhance the quality of {\KG} by ranking and filtering medical images. Both expert qualitative evaluations and quantitative benchmarking validate that {\KG} achieves high quality and is well-suited for practical downstream use.

To address \textbf{C2} (\textit{Utility Concern}), we conduct extensive experiments across two complementary settings to demonstrate the practical utility of {\KG}, as shown in Figure~\ref{fig:intro}. First, in the setting of knowledge graph analysis, we assess the intrinsic quality of {\KG} through a link prediction task. Second, in the setting of knowledge graph augmentation, we integrate {\KG} into downstream applications, including medical text-image retrieval and visual question answering. Our comprehensive evaluation spans 24 baselines, 4 vision-language backbones, and 6 datasets covering 5 distinct tasks. This broad evaluation framework allows us to systematically explore how {\KG} contributes to downstream performance. From these experiments, we derive several key insights:
\begin{itemize}[leftmargin=*]
\item \textit{Model Choice Should Align with Graph Structure}: Effective modeling of multimodal medical knowledge graphs requires selecting well-suited network architectures to handle their heterogeneous and relational nature, underscoring the importance of matching model design to graph characteristics.

\item \textit{External Knowledge Improves Downstream Tasks}: Incorporating structured medical knowledge consistently enhances downstream applications such as image–text retrieval and visual question answering, though the extent of improvement depends on the integration strategy and the underlying model architecture.

\item \textit{Balancing Knowledge Integration and Model Robustness}: While external knowledge generally improves coverage and reasoning capability, it also introduces challenges related to precision, recall and overfitting, highlighting the need for selective and adaptive knowledge fusion mechanisms.

\item \textit{Future Work Needs Unified and Adaptive Frameworks}: Advancing the field will require developing integration strategies that are both backbone-agnostic and adaptable, enabling knowledge graphs to be leveraged effectively across pretraining and fine-tuning stages for robust, generalizable improvements.
\end{itemize}

In summary, our contributions are threefold:
\begin{itemize}[leftmargin=*]
\item \textbf{Construction of {\KG}}: We present {\KG}, a new medical multimodal knowledge graph that integrates clinical terminology and visual instances, providing a crucial resource for the development of knowledge-intensive multimodal models.
\item \textbf{Effective Multimodal Knowledge Graph Filtering Algorithm}: We introduce Neighbor-aware Filtering (NaF), a targeted metric for ranking and filtering images in the context of a multimodal knowledge graph, which helps maintain the graph’s quality and conciseness.
\item \textbf{Extensive Benchmarking}: We conduct comprehensive evaluations spanning 5 tasks, 2 experimental settings, 24 baseline methods, 4 vision-language backbones, and 6 diverse datasets. Our results demonstrate that {\KG} meaningfully improves performance on knowledge-intensive medical applications and opens the door to new adaptive fusion strategies in multimodal learning.
\end{itemize}

\section{Related Work}

\textbf{Multimodal Learning in the Medical Domain.}
Multimodal learning has been widely applied to medical tasks such as criticality prediction~\cite{zhong2025medditpro, wang2024unity, xu2018raim, zhong2024synthesizing, feng2019dcmn, tang2020democratizing}, readmission prediction~\cite{yang2021leverage, wang2023hierarchical, wang2024unity}, adverse drug reaction prediction~\cite{luo2023padr}, and medical visual question answering~\cite{li2024llava, moor2023med, wang2024knowledge, wang2024fedkim, wangfedmeki}. However, most existing methods are largely data-driven and rely on task-specific datasets, with limited use of explicit structured knowledge. This limitation reduces their effectiveness on knowledge-intensive problems and motivates the development of more knowledge-reliable methods and benchmarks.

\smallskip
\noindent\textbf{Medical Knowledge Graphs.}
Medical knowledge graphs play an important role in organizing biomedical knowledge and facilitating downstream techniques such as representation learning~\cite{bordes2013translating,balavzevic2019hypernetwork,sun2019rotate} and retrieval-augmented generation~\cite{wang2026mkg,wang2026gpr,wang2026kamr}. Traditional resources have supported both broad biomedical systems~\cite{donnelly2006snomed, bodenreider2004unified, lipscomb2000medical} and specialized domains~\cite{wishart2006drugbank, goh2007human}, but they typically require extensive manual curation and expert involvement. These challenges become even greater when extending to multimodal medical data, especially imaging. To improve scalability, prior works have explored automated construction of medical knowledge graphs in comprehensive~\cite{lin2015learning, chandak2023building} and domain-specific settings, including pharmacology~\cite{bean2017knowledge, zhang2021prediction, wang2021adverse}, broader biomedicine~\cite{vlietstra2017automated, fei2021enriching, yuan2020constructing}, and COVID-19~\cite{michel2020covid}. However, many automated approaches still rely on simplified or outdated techniques, which may limit graph quality.

\smallskip
\noindent\textbf{Multimodal Knowledge Graphs.}
Recent work has extended unimodal knowledge graphs to multimodal settings. Existing approaches commonly construct multimodal knowledge graphs using search engines~\cite{wang2020richpedia, zhang2022multimodal, liu2019mmkg}, web crawlers~\cite{wang2023tiva, onoro2017answering}, or open knowledge sources such as Wikipedia~\cite{wang2020richpedia, zhang2022multimodal}. While these methods are often effective in general domains, their retrieval inaccuracies can substantially degrade graph quality in the medical domain, where high precision and reliability are essential.
\section{Construction of {\KG}}
\label{sec:construction}

\subsection{Problem Formulation}

Constructing a multimodal radiological knowledge graph from scratch poses significant challenges due to the scale, complexity, and heterogeneity of data modalities. A more practical and reliable strategy is to extend an existing unimodal knowledge graph by systematically incorporating additional modalities. In this work, we formulate the construction of our multimodal radiological knowledge graph as a \textit{modality-wise graph extension} problem.

We begin with the Unified Medical Language System (UMLS)~\cite{bodenreider2004unified}, a comprehensive biomedical knowledge base that standardizes and interconnects diverse health-related vocabularies via concept unique identifiers (CUIs). UMLS offers a rich repository of medical concepts and semantic relationships, serving as the foundational backbone for structured medical knowledge integration. For example, the clinical relation ``\textit{aspirin is used to treat myocardial infarction}'' is represented as a triplet (C0011849, treats, C0020538), where ``{C0011849}'' corresponds to ``\textit{Aspirin}'' and ``{C0020538}'' to ``\textit{Myocardial Infarction (Heart Attack)}''.

We expand the UMLS graph by introducing radiological image nodes and establishing cross-modal edges. The resulting graph contains two types of nodes: (1) \textbf{clinical concepts}, inherited directly from UMLS, and (2) \textbf{radiological images}. It also includes two types of edges: (1) \textbf{intra-modality edges} among clinical concepts (as defined in UMLS), and (2) \textbf{cross-modality edges} that link clinical concepts to corresponding images.

To perform the multimodal extension, we leverage the MIMIC-CXR dataset~\cite{johnson2019mimic}, which consists of paired radiology reports and chest X-ray images.
% \footnote{Details about the preprocessing of MIMIC-CXR are available in \appref{app:preprocessing} at \url{https://github.com/XiaochenWang-PSU/MedMKG/blob/main/Appendix.pdf}.}
From each report, we extract relevant clinical concepts and align them with their associated images, thereby establishing meaningful cross-modal connections. This design enables the extended knowledge graph to seamlessly integrate textual and visual information within a unified and structured framework. 
% The pipeline of constructing ~\KG  is illustrated in Figure~\ref{fig:system}, while more details are further elaborated in the following subsections. 

% \begin{figure}[h]
%     \centering
%     \includegraphics[width=0.99\linewidth]{fig/system-cropped.pdf}
%     \caption{Pipeline of constructing {\KG}.}
%     \label{fig:system}
% \end{figure}

\subsection{Concept Extraction}

A central challenge in constructing {\KG} lies in accurately establishing cross-modal edges between radiological images and clinical concepts. To address this, we design a two-stage pipeline that leverages the complementary strengths of rule-based systems and large language models (LLMs). Rule-based tools are highly effective in handling clinical terminologies and ontologies, offering broad coverage of domain-specific entities. In contrast, LLMs provide strong contextual understanding and disambiguation capabilities, enabling more accurate interpretation of report-level semantics. By integrating these two approaches, our pipeline achieves both the comprehensive coverage and semantic precision necessary for high-quality concept extraction and reliable cross-modal alignment.

\textbf{Stage I – Concept Identification.}
We begin by applying MetaMap~\cite{aronson2010overview}, a widely used rule-based tool, to each radiology report to identify candidate mentions of UMLS concepts. This step produces an exhaustive set of potential concept mappings for each mention, ensuring comprehensive coverage of clinically relevant entities. To focus on concepts with clinical significance, we filter out irrelevant semantic types based on domain knowledge.
% \footnote{A complete list of excluded semantic types is provided in \appref{app:semantic_type}.}

\textbf{Stage II – Concept Disambiguation.}
Next, we refine the candidate concepts using ChatGPT-4o~\cite{openai2023gpt} that considers both the full radiology report and the list of extracted candidates. For each mention, the LLM is prompted to select the most contextually appropriate concept, leveraging its strong semantic understanding to resolve ambiguity. This stage eliminates spurious or out-of-context candidates, resulting in a clean and accurate set of disambiguated clinical concepts aligned with each image.

This two-stage design enables precise and context-aware mapping of clinical concepts to radiological images, ensuring the construction of high-quality cross-modal edges in the resulting knowledge graph. Aggregating the selected concepts across all mentions in a report yields the final set of clinical concepts associated with each image.

\subsection{Relation Extraction}

With the clinical concepts identified, we further enrich the knowledge graph by establishing relations:

\textbf{Intra-Modality Relations.}
We introduce edges between identified clinical concepts whenever a relation is defined between them in UMLS. Only validated relations connecting distinct concepts are added, ensuring that intra-modality relationships are medically accurate and standardized.

\textbf{Cross-Modality Relations.}
Each image is linked to its extracted clinical concepts through cross-modality edges. However, beyond simply linking images and concepts, we also assign a semantic label to each edge to reflect the nature of the relationship. Specifically, each relation is categorized as \textit{Positive}, \textit{Negative}, or \textit{Uncertain}, indicating whether the concept is supported by, contradicted by, or ambiguously discussed in the report.

While the intra-modality relations are extracted through querying the UMLS knowledge base, the cross-modal relation extraction is performed jointly with concept disambiguation. During the LLM prompting process, the model is additionally instructed to assess the semantic stance (positive, negative, or uncertain) between the image and each concept. These relation labels are used to annotate the edges accordingly.
% \footnote{ Details concerning the prompting procedure are available in \appref{app:prompt}, while the analysis on the selection of LLM can be found in \appref{app:llm}.}

% \subsection{Informativeness-Oriented Filtering}

\begin{table}[t]
\centering\small
\caption{Data Statistics Summary}
\label{tab:data_stats}
\vspace{-0.15in}
% \vspace{2mm}
\resizebox{0.65\linewidth}{!}{%
\begin{tabular}{l r}
\toprule
\textbf{Statistic} & \textbf{Count} \\
\midrule
Total Number of Edges          & 35,387 \\
Number of Concepts        & 3,149    \\
Number of Images          & 4,868   \\
Number of Relations       & 262       \\
Number of Cross-modality Edges  & 20,705  \\
Number of Intra-modality Edges & 14,682 \\
Image-to-Concept Ratio           & 1.55      \\
Average Edges per Image        & 4.25     \\
Average Edges per Concept      & 11.24    \\
\bottomrule
\end{tabular}
}
\vspace{-0.1in}
\end{table}

\subsection{Neighbor-Aware Filtering for Image Informativeness}
\label{sec:naf}
The full construction process produces a highly comprehensive multimodal knowledge graph. However, its large scale, with numerous images and associated concepts, creates challenges for storage, computation, and downstream analysis. In particular, many radiological images are \textit{redundant} because they capture similar and homogeneous regions~\cite{zhou2010redundancy}. This redundancy can overwhelm subsequent analysis and reduce graph efficiency. To improve efficiency without sacrificing knowledge quality, we introduce a filtering strategy that prioritizes the most informative and distinctive images.

Ideally, a representative medical image should be connected to multiple clinical concepts through diverse relations, making the number of its neighboring nodes a key indicator of informativeness. However, relying solely on the number of neighbors may introduce noise, as some medical concepts are linked to a large number of generic or non-discriminative images. To mitigate this, we additionally consider the distinctiveness of an image in the context of its 2-hop neighborhood. Intuitively, if a relation–concept pair is associated with only a few images, those images are likely to carry more unique and clinically informative content.

Based on this insight, we propose a \textbf{Neighbor-aware Filtering (NaF)} strategy that balances both connectivity and distinctiveness. The informativeness score of an image $m$ is defined as:
% \begin{equation}
%     \text{Info-Score}(m) = \sum_{(r,c)\in \mathcal{N}_{m}}\log \frac{M}{|\mathcal{N}_{(r,c)}|},
% \end{equation}
\begin{equation}
    \text{NaF}(m) = \sum_{(r,c)\in \mathcal{N}_{m}}\log \frac{M}{|\mathcal{N}_{(r,c)}|},
\end{equation}
where each triplet $(m,r,c)$ represents a connection between image $m$, relation $r$, and concept $c$; $\mathcal{N}_{m}$ denotes the 1-hop neighbors of $m$; $M$ represents the number of medical images in the knowledge graph; and $\mathcal{N}_{(r,c)}$ is the set of images linked to $c$ via $r$.

By combining these two dimensions, the designed NaF strategy effectively prioritizes images that are both rich in clinical content and contribute unique, informative knowledge to the graph. After computing the informativeness scores, we rank all images in descending order and select them from top to bottom until the full set of concepts is covered. This strategy ensures that the final graph retains maximal clinical richness and diversity while eliminating redundant or overly generic images, thereby improving scalability and downstream utility.

\subsection{Quantitative and Qualitative Analysis}

% \begin{wraptable}{r}{0.45\columnwidth}
% \centering
% \caption{Data Statistics Summary}
% \label{tab:data_stats}
% \resizebox{\linewidth}{!}{%
% \begin{tabular}{l r}
% \toprule
% \textbf{Statistic} & \textbf{Value} \\
% \midrule
% Total Number of Edges          & 35,387 \\
% Number of Concepts        & 3,149    \\
% Number of Images          & 4,868   \\
% Number of Relations       & 262       \\
% Number of Cross-modality Edges  & 20,705  \\
% Number of Intra-modality Edges & 14,682 \\
% Image-to-Concept Ratio           & 1.55      \\
% Average Edges per Image        & 4.25     \\
% Average Edges per Concept      & 11.24    \\
% \bottomrule
% \end{tabular}
% }
% \end{wraptable}

% \begin{wrapfigure}{r}{0.45\columnwidth}
%     \centering
%         \caption{Human assessment results for {\KG}. The scores are on a scale from 0 to 10.}\includegraphics[width=0.85\linewidth]{fig/evaluation_boxplot.pdf}

%     \label{fig:human_assessment}
% \end{wrapfigure}

To acquire an intuitive understanding of {\KG}'s statistical characteristics and soundness of {\KG }, we performed both quantitative and qualitative analyses.

\textbf{Quantitative Analysis.}
{\KG}'s statistics are detailed in Table~\ref{tab:data_stats}.
% {\KG} consists of 4,868 images, 3,149 clinical concepts, and 35,387 edges, as summarized in Table~\ref{tab:data_stats}. Approximately 20,700 of these edges link images to concepts, while the remainder capture relationships between clinical concepts.
The moderate scale of {\KG} facilitates convenient utilization in diverse application scenarios with different computational budgets. Additionally, images and concepts are intensively connected with intra- and cross-modal neighbors, promoting rich multimodal reasoning. 
% Furthermore, Figure~\ref{fig:semantic_type} in \appref{app:overview} shows the distribution of semantic types between the clinical concepts involved, indicating a broad and balanced coverage of the areas of clinical knowledge.

% \begin{table}[h!]
% \centering
% \caption{Data Statistics Summary}
% \label{tab:data_stats}
% \resizebox{0.5\columnwidth}{!}{%
% \begin{tabular}{l r}
% \toprule
% \textbf{Statistic} & \textbf{Value} \\
% \midrule
% Total Number of Edges          & 35,387 \\
% Number of Concepts        & 3,149    \\
% Number of Images          & 4,868   \\
% Number of Relations       & 262       \\
% Number of Cross-modality Edges  & 20,705  \\
% Number of Intra-modality Edges & 14,682 \\
% Image-to-Concept Ratio           & 1.55      \\
% Average Edges per Image        & 4.25     \\
% Average Edges per Concept      & 11.24    \\
% \bottomrule
% \end{tabular}
% }
% \end{table}

% \setlength{\tabcolsep}{2.5pt}

\begin{table*}[t!]
\centering
\caption{Performance of 17 approaches on three link prediction tasks. \textbf{Bold}
indicates the best value among baselines.}
\vspace{-0.15in}
\resizebox{.95\textwidth}{!}{%
\begin{tabular}{l|cccc|cccc|cccc}
\toprule
\multirow{2}{*}{\textbf{Model}} & \multicolumn{4}{c|}{\textbf{\textcolor{red}{Head Prediction}}} & \multicolumn{4}{c|}{\textbf{\textcolor{blue}{Relation Prediction}}} & \multicolumn{4}{c}{\textbf{\textcolor{violet}{Tail Prediction}}} \\ \cline{2-13}
& \textbf{MR} $\downarrow$ & \textbf{Hits@3} $\uparrow$ & \textbf{Hits@5} $\uparrow$ & \textbf{Hits@10} $\uparrow$
& \textbf{MR} $\downarrow$ & \textbf{Hits@3} $\uparrow$ & \textbf{Hits@5} $\uparrow$ & \textbf{Hits@10} $\uparrow$
& \textbf{MR} $\downarrow$ & \textbf{Hits@3} $\uparrow$ & \textbf{Hits@5} $\uparrow$ & \textbf{Hits@10} $\uparrow$ \\
\midrule
TransR   & 1505.66 & 1.80 & 3.71 & 7.50 & 106.01 & 5.84 & 10.55 & 19.98 & 887.07 & 3.34 & 6.65 & 13.11 \\
TransD   & \textbf{1219.92} & \textbf{3.84} & \textbf{7.56} & \textbf{12.22} & 53.49 & 27.49 & 35.52 & 46.36 & 586.61 & 5.25 & \textbf{10.90} & \textbf{18.67} \\
TransE   & 1248.36 & 3.36 & 6.33 & 9.93 & 39.74 & 20.13 & 28.79 & 42.45 & 544.79 & 4.99 & 9.20 & 15.14 \\
TransH   & 1263.25 & 3.29 & 6.16 & 9.99 & 37.40 & 22.02 & 30.21 & 43.41 & 561.39 & 5.05 & 9.67 & 16.07 \\
RotatE   & 1560.49 & 1.35 & 2.69 & 5.04 & 129.84 & 0.77 & 1.42 & 3.25 & 739.95 & 1.54 & 3.47 & 6.82 \\
\hline
DistMult & 3590.63 & 0.10 & 0.19 & 0.38 & 119.13 & 2.17 & 3.92 & 6.95 & 3582.56 & 0.15 & 0.26 & 0.50 \\
SimplE   & 4032.57 & 0.01 & 0.04 & 0.10 & 133.16 & 0.63 & 1.19 & 2.95 & 4033.27 & 0.05 & 0.06 & 0.11 \\
TuckER   & 1533.74 & 2.83 & 4.51 & 7.45 & 43.75 & \textbf{46.67} & \textbf{55.42} & \textbf{64.01} & 1235.88 & 4.31 & 7.12 & 11.71 \\
ComplEx  & 3790.54 & 0.09 & 0.20 & 0.31 & 125.58 & 0.93 & 1.81 & 3.79 & 3782.86 & 0.12 & 0.23 & 0.40 \\
RESCAL   & 3849.47 & 0.03 & 0.06 & 0.12 & 127.51 & 0.44 & 0.99 & 2.40 & 3845.42 & 0.02 & 0.03 & 0.07 \\
\hline
HypER    & 3564.84 & 0.33 & 0.56 & 0.93 & 122.79 & 1.16 & 2.20 & 4.12 & 2933.17 & 1.07 & 1.71 & 2.89 \\
ConvE    & 2071.83 & 1.81 & 2.76 & 4.59 & 59.35 & 18.79 & 26.51 & 36.10 & 777.91 & 4.28 & 6.87 & 11.03 \\
ConvR    & 2438.14 & 0.62 & 1.05 & 1.77 & 78.55 & 6.12 & 10.17 & 17.75 & 787.58 & 2.25 & 3.80 & 6.83 \\
\hline
AttH     & 2113.85 & 0.36 & 0.86 & 1.66 & \textbf{31.69} & 26.94 & 36.91 & 50.45 & \textbf{523.86} & \textbf{5.72} & 8.90 & 14.10 \\
MurE & 1266.26 & 3.62 & 6.22 & 9.63 & 40.50 & 19.27 & 27.71 & 41.38 & 567.77 & 4.92 & 8.67 & 15.79\\
MurP     & 2771.43 & 2.14 & 4.29 & 7.00 & 158.56 & 3.45 & 4.67 & 7.02 & 590.11 & 4.34 & 7.59 & 12.64 \\
\hline
NTN      & 4007.11 & 0.01 & 0.02 & 0.06 & 140.38 & 0.14 & 0.29 & 1.02 & 3994.01 & 0.01 & 0.02 & 0.05 \\
\bottomrule
\end{tabular}%
}
\vspace{-0.05in}
\label{tab:kg_baselines}
\end{table*}

% \subsection{Assessment}

\begin{figure}[t]
    \centering
    
    % \vspace{-0.05in}
    \includegraphics[width=0.65\linewidth]{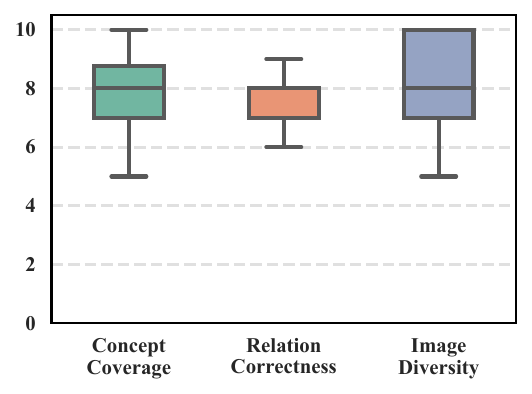}
    \vspace{-0.20in}
    \caption{Human assessment results.}
    \label{fig:demo}
    \vspace{-0.15in}
\end{figure}
\textbf{Qualitative Analysis.}
To further assess the quality of {\KG}, we conducted a human evaluation with experienced radiologists. The experts reviewed a set of sampled subgraphs and assigned quality scores across three key dimensions, each rated on a scale from 1 to 10: (1) \textit{concept coverage} — whether the graph captures the key image-related clinical concepts; (2) \textit{relation correctness} — whether the cross-modal relations are accurately identified; and (3) \textit{image diversity} — whether the linked images reflect a broad range of clinical scenarios. Higher scores indicate better performance on each metric. {\KG} achieves an average score of 7.789/10 (77.89\%) across all three metrics. Compared with previous studies where 60\% of agreement is regarded convincing~\cite{schafer2024biokgrapher, kilicoglu2008semantic}, this result indicates {\KG}'s reliability and practical utility as a multimodal medical knowledge source.
% \footnote{An illustration of the constructed {\KG} is available in \appref{app:overview}. Further details on the evaluation protocol are provided in \appref{app:human_asssessment}. License information is available in \appref{app:license}.}

% \begin{wrapfigure}{r}{0.45\columnwidth}
%     \centering
%         \caption{Human assessment results for {\KG}. The scores are on a scale from 0 to 10.}\includegraphics[width=0.85\linewidth]{fig/evaluation_boxplot.pdf}

%     \label{fig:human_assessment}
% \end{wrapfigure}

\section{Benchmark}
\label{sec:benchmark}

\begin{table*}[t]
\centering
\caption{Results (\%) on text-image retrieval task for OpenI and MIMIC-CXR datasets. Notable augmentation with {\KG} is observed, especially for KnowledgeCLIP.}
\vspace{-0.15in}
\label{tab:it_retrieval_reduced}
\resizebox{0.77\textwidth}{!}
{
\begin{tabular}{l|ccc|ccc|ccc|ccc}
\toprule
\multirow{3}{*}{\textbf{Methods}} & \multicolumn{6}{c|}{\textbf{OpenI}} & \multicolumn{6}{c}{\textbf{MIMIC-CXR}} \\ \cline{2-13}
 & \multicolumn{3}{c|}{\textbf{Precision @$K$ 
 $\uparrow$}} & \multicolumn{3}{c|}{\textbf{Recall @$K$ $\uparrow$}} & \multicolumn{3}{c|}{\textbf{Precision @$K$ $\uparrow$}} & \multicolumn{3}{c}{\textbf{Recall @$K$ $\uparrow$}} \\ \cline{2-13}
 & 10 & 20 & 100 & 10 & 20 & 100 & 10 & 20 & 100 & 10 & 20 & 100 \\ \hline
\textbf{CLIP} & 1.17 & 1.00 & 0.56 & 11.10 & 19.24 & 53.48  & 1.11 & 0.98 & 0.58 & 11.11 & 19.52 & 58.26   \\
$+$ FashionKLIP & 1.29 & 1.16 & 0.63 & 12.64 & 22.75 & 60.46  & 1.19 & 0.99 & 0.56 & 11.91 & 19.82 & 56.06 \\
$+$ KnowledgeCLIP & \textbf{2.63} & \textbf{1.99} & \textbf{0.79} & \textbf{25.56} & \textbf{38.83} & \textbf{76.16} & \textbf{2.33} & \textbf{1.73} & \textbf{0.74} & \textbf{23.32} & \textbf{34.53} & \textbf{74.37}  \\\hline
\textbf{PubMedCLIP} & 1.17 & 0.98 & 0.51 & 10.81 & 18.47 & 48.46 & 0.69 & 0.65 & 0.43 & 6.91 & 13.01 & 42.79   \\
$+$ FashionKLIP  & \textbf{1.54} & \textbf{1.21} & \textbf{0.70} & \textbf{15.10} & \textbf{23.38} & \textbf{67.73} & 0.73 & 0.72 & 0.49 & 7.31 & 14.41 & 49.20\\
$+$ KnowledgeCLIP  & 1.49 & 1.17 & 0.61 & 14.33 & 22.61 & 59.41 & \textbf{1.26} & \textbf{1.13} & \textbf{0.60} & \textbf{12.61} & \textbf{22.62} & \textbf{59.96}  \\\hline
\textbf{BiomedCLIP} & 1.04 & 0.79 & 0.42 & 9.90 & 15.10 & 40.45 & 2.02 & 1.59 & 0.66 & 20.12 & 31.63 & 65.77  \\
$+$ FashionKLIP  & \textbf{1.46} & \textbf{1.15} & \textbf{0.60} & \textbf{14.33} & \textbf{22.47} & \textbf{58.22}  & 2.02 & 1.49 & 0.68 & 20.12 & 29.63 & 67.77 \\
$+$ KnowledgeCLIP  & 1.26 & 0.95 & 0.49 & 12.50 & 18.61 & 47.54 & \textbf{2.64} & \textbf{1.94} & \textbf{0.71} & \textbf{26.33} & \textbf{38.74} & \textbf{70.77}  \\ \hline
\textbf{MedCSPCLIP} & 1.60 & 1.10 & 0.54 & 15.73 & 21.35 & 52.14 & 3.77 & 2.59 & 0.82 & 37.69 & 51.65 & 81.58 \\
$+$ FashionKLIP & 1.81 & 1.36 & 0.60 & 17.84 & 26.54 & 57.65   & 4.02 & 2.69 & 0.85 & 40.19 & 53.75 & 84.98  \\
$+$ KnowledgeCLIP & \textbf{1.90} & \textbf{1.40} & \textbf{0.62} & \textbf{18.61} & \textbf{27.18} & \textbf{59.55} & \textbf{4.95} & \textbf{3.14} & \textbf{0.89} & \textbf{49.50} & \textbf{62.66} & \textbf{88.99}  \\
\bottomrule
\end{tabular}
}
\vspace{-0.05in}
\end{table*}

\begin{table*}[ht]
\centering
\caption{Results (\%) on knowledge-augmented visual question answering. Best results within each backbone block are in \textbf{bold}.}
\vspace{-0.1in}
\label{tab:vqa_baselines}
\resizebox{0.77\textwidth}{!}{%
\begin{tabular}{l|cccc|cccc|cccc}
\toprule
\multirow{2}{*}{\textbf{Methods}} & \multicolumn{4}{c|}{\textbf{VQA-RAD}} & \multicolumn{4}{c|}{\textbf{SLAKE}} & \multicolumn{4}{c}{\textbf{PathVQA}} \\
\cmidrule(lr){2-13}
 & Acc$\uparrow$ & Prec$\uparrow$ & Rec$\uparrow$ & F1$\uparrow$ & Acc$\uparrow$ & Prec$\uparrow$ & Rec$\uparrow$ & F1$\uparrow$ & Acc$\uparrow$ & Prec$\uparrow$ & Rec$\uparrow$ & F1$\uparrow$ \\
\midrule

\textbf{CLIP} & 64.94 & 62.71 & 62.71 & 62.71 & 65.07 & 62.09 & \textbf{74.86} & 67.88 & 81.89 & 88.37 & 76.54 & 82.03 \\
$+$ KRISP & \textbf{73.71} & \textbf{78.89} & 60.17 & 68.27 & 56.90 & 55.00 & 69.14 & 61.27 & 84.21 & \textbf{89.83} & 79.79 & 84.51 \\
$+$ MKBN & 70.12 & 70.87 & 61.86 & 66.06 & 70.14 & 73.47 & 61.71 & 67.08 & 84.68 & 89.35 & 81.33 & 85.15 \\
$+$ K-PathVQA & 66.14 & 62.79 & \textbf{68.64} & 65.59 & 69.30 & 73.57 & 58.86 & 65.40 & 84.15 & 85.74 & 84.75 & 85.24 \\
$+$ EKGRL & 67.73 & 65.04 & 67.80 & 66.39 & 70.70 & 71.01 & 68.57 & 69.77 & \textbf{84.77} & 86.38 & 85.24 & \textbf{85.81} \\
$+$ MR-MKG & \textbf{73.71} & 77.08 & 62.71 & \textbf{69.16} & \textbf{76.34} & \textbf{79.74} & 69.71 & \textbf{74.39} & 84.30 & 84.85 & \textbf{86.34} & 85.59 \\
\midrule

\textbf{BioMedCLIP} & 66.93 & 61.74 & 77.97 & 68.91 & 70.14 & 70.18 & 68.57 & 69.36 & 84.56 & \textbf{94.03} & 76.27 & 84.22 \\
$+$ KRISP & 76.10 & \textbf{79.59} & 66.10 & 72.22 & 57.18 & 54.77 & 75.43 & 63.46 & 85.46 & 93.74 & 78.30 & 85.33 \\
$+$ MKBN & 68.53 & 64.66 & 72.88 & 68.53 & 67.89 & 75.63 & 51.43 & 61.22 & 85.78 & 88.32 & 84.91 & 86.58 \\
$+$ K-PathVQA & 65.34 & 71.23 & 44.07 & 54.45 & 70.70 & 73.20 & 64.00 & 68.29 & 85.93 & 90.57 & 82.54 & 86.37 \\
$+$ EKGRL & 75.70 & 71.76 & \textbf{79.66} & 75.50 & \textbf{86.20} & \textbf{89.38} & \textbf{81.71} & \textbf{85.37} & 85.46 & 89.66 & 82.60 & 85.98 \\
$+$ MR-MKG & \textbf{77.29} & 74.80 & 77.97 & \textbf{76.35} & 80.28 & 79.66 & 80.57 & 80.11 & \textbf{87.24} & 90.06 & \textbf{85.85} & \textbf{87.91} \\
% \midrule

% \textbf{MedCSPCLIP} & 68.13 & 61.59 & \textbf{85.59} & 71.63 & 66.20 & 83.95 & 38.86 & 53.12 & 77.72 & 73.37 & \textbf{92.24} & 81.73 \\
% $+$ KRISP & \textbf{80.08} & \textbf{84.00} & 71.19 & \textbf{77.06} & 70.70 & \textbf{91.76} & 44.57 & 60.00 & 83.19 & \textbf{94.71} & 72.96 & 82.43 \\
% $+$ MKBN & 69.72 & 65.44 & 75.42 & 70.08 & 67.32 & 75.21 & 50.29 & 60.27 & 85.37 & 86.17 & 86.84 & 86.51 \\
% $+$ K-PathVQA & 67.73 & 75.34 & 46.61 & 57.59 & 71.55 & 74.03 & 65.14 & 69.30 & 85.31 & 89.35 & 82.65 & 85.87 \\
% $+$ EKGRL & 76.10 & 73.39 & 77.12 & 75.21 & 69.30 & 78.95 & 51.43 & 62.28 & 84.92 & 92.75 & 78.19 & 84.85 \\
% $+$ MR-MKG & 78.49 & 77.59 & 76.27 & 76.92 & \textbf{83.94} & 83.15 & \textbf{84.57} & \textbf{83.85} & \textbf{86.53} & 89.74 & 84.75 & \textbf{87.17} \\
\bottomrule
\end{tabular}%
}
\vspace{-0.05in}
\end{table*}

We evaluate {\KG} under two complementary scenarios: \textbf{knowledge graph analysis} and \textbf{knowledge graph augmentation}. In the knowledge graph analysis setting, we assess tasks that directly utilize the internal structure and semantics of the graph, i.e., \textit{link prediction}.
In the knowledge graph augmentation setting, {\KG} is employed as auxiliary knowledge to enhance the performance of external multimodal applications, including \textit{multimodal text-image retrieval} and \textit{multimodal visual question answering (VQA)}. 

\subsection{Link Prediction}
\label{sec:link_prediction}

The link prediction tasks~\cite{bordes2013translating} focus on inferring missing links between entities by predicting either the head entity, the tail entity, or the relation connecting them. Specifically, given two known components of a triple, such as a relation and one entity, or two entities, the goal is to predict the missing element that completes the triple. This task helps improve the completeness and utility of knowledge graphs by filling in missing entities or relations between entities.
\noindent \textbf{Evaluation Results.}
Table~\ref{tab:kg_baselines} compares 17 link prediction baselines on head, relation, and tail prediction over our {\KG}. A clear gap appears between head and tail prediction: models perform better on tail entities, which are exclusively clinical concepts, than on head entities, which include both images and concepts. This heterogeneity makes head prediction more difficult because it requires aligning multimodal representations in a shared embedding space.
Among all baselines, translation-based models (TransD, TransE, TransH) achieve the strongest overall performance. Tensor factorization models show mixed results: TuckER performs relatively well on relation prediction, while others such as SimplE and RESCAL remain weak, suggesting limited consistency in entity linking.
Overall, these results highlight the need for link prediction methods that better match the multimodal and relational structure of medical knowledge graphs. Future work may explore hybrid designs that combine the strengths of translation-based and factorization-based models, as well as modality-aware methods to improve head prediction.

\subsection{Knowledge-Augmented Text-Image Retrieval}
\label{sec:retrieval}

The knowledge-augmented text-image retrieval task aims to enhance conventional medical text-image retrieval~\cite{demner2012design} by leveraging domain knowledge encoded in a multimodal medical KG.

\textbf{Evaluation Results.}
Table~\ref{tab:it_retrieval_reduced} shows that knowledge augmentation generally improves retrieval performance across both OpenI and MIMIC-CXR, particularly in low-K settings. This indicates that external knowledge enhances the model’s ability to identify the most relevant matches at top ranks. Among the two strategies, KnowledgeCLIP (postraining-based) shows strong and consistent gains across most settings, especially on MIMIC-CXR, while FashionKLIP (joint fine-tuning) provides more noticeable improvements on OpenI relative to its effect on MIMIC-CXR.
The overall trend suggests that integrating external knowledge, whether through pretraining or joint fine-tuning, can significantly benefit medical retrieval tasks. 
% However, the effectiveness varies with the backbone and integration method, underscoring the importance of alignment between knowledge signals and visual-language representations. 
Future work may explore tighter coupling between knowledge and model training by involving the medical knowledge graph in both pretraining and fine-tuning stages. Such unified frameworks could offer deeper semantic grounding and more robust generalization across diverse clinical retrieval scenarios.

\subsection{Knowledge-Augmented Visual Question Answering}
\label{sec:vqa}

The knowledge-augmented visual question answering task aims to improve medical visual question answering task~\cite{hasan2018overview} by integrating domain knowledge contained in multimodal medical knowledge graphs, enabling more accurate and clinically meaningful question answering over medical images.
\noindent\textbf{Evaluation Results.}
Table~\ref{tab:vqa_baselines} reports the performance (\%) of knowledge-augmented VQA models on VQA-RAD, SLAKE, and PathVQA. Incorporating external knowledge from our multimodal medical knowledge graph generally improves Accuracy and F1, demonstrating the value of structured domain-specific knowledge for medical visual reasoning.
Among the compared methods, MR-MKG achieves the strongest and most stable results across datasets and backbones, highlighting the effectiveness of contrastive learning for cross-modal alignment. In contrast, attention-based fusion methods (K-PathVQA and MKBN) show less consistent gains and tend to degrade on smaller datasets such as VQA-RAD and SLAKE, likely due to overfitting, while becoming more stable on the larger PathVQA dataset.
Overall, these results confirm the effectiveness of our multimodal medical knowledge graph in improving medical VQA. Its clinical specificity, image-aware relational structure, and semantic richness support stronger multimodal understanding. Future work can further explore more adaptive and backbone-agnostic fusion strategies.

\subsection{Ablation on NaF}

To understand how NaF improves the utility of {\KG}, we conduct an ablation study using KRISP on the PathVQA dataset with three versions of the graph: (i) the graph filtered by our proposed NaF algorithm, (ii) a graph obtained via random sampling to match NaF’s size, and (iii) an unfiltered graph. The results are shown in Table~\ref{tab:ablation_naf}.
The model that relies on the unfiltered graph struggles to extract useful signals due to severe redundancy and noise. As a result, it tends to adopt an overly conservative prediction strategy, yielding high precision but substantially worse recall, and ultimately performs poorly on overall accuracy and F1. Random filtering, by contrast, reduces redundancy and helps the model access more relevant information, but it also indiscriminately removes informative nodes and relations, degrading graph quality and leading to suboptimal performance.
\begin{table}[t]
\centering
\caption{Ablation study results.}
\vspace{-0.15in}
\label{tab:ablation_naf}
\resizebox{\columnwidth}{!}{%
\begin{tabular}{lcccc}
\toprule
\textbf{Ablation} & \textbf{Accuracy} & \textbf{Precision} & \textbf{Recall} & \textbf{F1} \\
\midrule
{\KG} w.\ NaF              & \textbf{84.2} & 89.8 & \textbf{79.8} & \textbf{84.5} \\
{\KG} w.\ Random Filtering & 80.7 & 96.9 & 66.5 & 78.4 \\
{\KG} w/o Filtering        & 79.4 & \textbf{98.1} & 62.6 & 76.4 \\
\bottomrule
\end{tabular}%
}
\vspace{-0.06in}
\end{table}
NaF achieves the best results among all three settings. By selectively removing redundant structure while preserving essential graph information, NaF provides a cleaner, more discriminative knowledge graph. This confirms NaF’s effectiveness in reducing structural redundancy and noise, consistent with our discussion in Section~\ref{sec:naf}.

% \begin{table*}[t!]
% \centering
% \caption{Ablation on NaF.}
% \vspace{-0.1in}
% \setlength{\tabcolsep}{3pt}
% \renewcommand{\arraystretch}{0.95}
% \resizebox{\textwidth}{!}{
% \begin{tabular}{l|cccc|cccc|cccc}
% \toprule
% \multirow{2}{*}{\textbf{Version}} & \multicolumn{4}{c|}{\textbf{\textcolor{red}{Head Prediction}}} & \multicolumn{4}{c|}{\textbf{\textcolor{blue}{Relation Prediction}}} & \multicolumn{4}{c}{\textbf{\textcolor{violet}{Tail Prediction}}} \\ \cline{2-13}
% & \textbf{MR} $\downarrow$ & \textbf{Hits@3} $\uparrow$ & \textbf{Hits@5} $\uparrow$ & \textbf{Hits@10} $\uparrow$ & \textbf{MR} $\downarrow$ & \textbf{Hits@3} $\uparrow$ & \textbf{Hits@5} $\uparrow$ & \textbf{Hits@10} $\uparrow$ & \textbf{MR} $\downarrow$ & \textbf{Hits@3} $\uparrow$ & \textbf{Hits@5} $\uparrow$ & \textbf{Hits@10} $\uparrow$ \\
% \midrule
% {\KG} & 1219.92 & 3.84 & 7.56 & 12.22 & 53.49 & 27.49 & 35.52 & 46.36 & 586.61 & 5.25 & 10.90 & 18.67 \\
% {\KG} w/o  NaF & 2064.32 & 12.70 & 13.87 & 15.64 & 79.88 & 17.65 & 22.81 & 31.09 & 356.61 & 15.79 & 19.90 & 26.65 \\
% \bottomrule
% \end{tabular}
% }
% \vspace{-0.12in}
% \label{tab:transd_ablation}
% \end{table*}

% \begin{figure}
%     \centering
%         \caption{The ablation study on the effectiveness of NaF.}
%         \vspace{-0.05in}
%     \includegraphics[width=1\linewidth]{fig/ablation_study_NaF.pdf}
%     \label{fig:ab}
%     \vspace{-0.3in}
% \end{figure}

% \input{tex/discussion}
\section{Conclusion}

In this work, we present {\KG}, a multimodal medical knowledge graph that integrates clinical text and medical imaging to capture rich intra- and cross-modality relationships. To improve graph quality and conciseness, we further introduce a neighbor-aware filtering algorithm tailored to multimodal settings. Extensive experiments on graph analysis and downstream augmentation tasks demonstrate the effectiveness of {\KG} and its value for medical knowledge representation. Beyond serving as an expandable resource, {\KG} also opens new directions for integrating multimodal knowledge into clinical applications such as report generation, diagnostic reasoning, and temporal prediction.

\clearpage
\section*{Acknowledgements}
This research was partially supported by a 2025/2026 Rising Researcher Grant from Penn State's Institute for Computational \& Data Sciences (RRID:SCR\_025154) and the National Science Foundation under Grant No. 2333790 and 2238275.

\bibliographystyle{ACM-Reference-Format}
% \balance
\bibliography{custom}

\appendix
\addtocontents{toc}{\protect\setcounter{tocdepth}{2}}
\clearpage

\section{Compute and Environment Configuration}\label{app:compute}
All experiments were conducted on an NVIDIA A100 GPU with CUDA version 12.0, running on an Ubuntu 20.04.6 LTS server. 
% Additional details regarding the experimental setup and environment can be found in the GitHub repository.

% \section{Dataset Repository}
% We have released a public dataset repository for {\KG}, available on Github at \textcolor{red}{\url{https://github.com/XiaochenWang-PSU/MedMKG}} and on Hugging Face at \textcolor{blue}{\url{https://huggingface.co/datasets/xcwangpsu/MedMKG}}. The {\KG} dataset can be loaded using the Hugging Face datasets module, alongside the MIMIC-CXR dataset, which requires separate download following the instructions provided in the Hugging Face repository README file. The GitHub repository includes runnable code for data processing, baseline models, environment configuration, and example execution scripts. We are committed to regularly updating the repository with additional modalities, datasets, and tasks to further support the research community.

\section{Dataset Repository}
We have provided a anonymous dataset repository for {\KG}, available at \textcolor{red}{\url{https://github.com/XiaochenWang-PSU/MedMKG/}}. The {\KG} dataset can be loaded alongside the MIMIC-CXR dataset, which requires separate download following the instructions provided in the repository README file. The repository also includes runnable code for data processing, baseline models, environment configuration, and example execution scripts. When using the data, users must adhere to the consent terms specified by the MIMIC-CXR maintainers.

% We are committed to publication of the repository after the acceptance of this study, as well as regularly updating the repository with additional modalities, datasets, and tasks to further support the research community.

\begin{figure*}[t]
\centering\small
\begin{tcolorbox}[
  title=Prompt for Concept Disambiguation and Relation Extraction,
  colback=white,
  colframe=blue!50!black,
  fonttitle=\bfseries,
  width=\textwidth,
  boxsep=1mm, left=1mm,right=1mm,top=1mm,bottom=1mm
]

\textbf{Report Text:} [\textit{Report Text}] \par
\vspace{0.5mm}
\textbf{Candidate Concepts:} [\textit{Candidate Concepts}] \par

\vspace{1mm}
For each phrase, evaluate the concept candidates and select the most relevant concept based on the report context. Your decision should account for the radiological image context.

\vspace{0.8mm}
After selecting the concept for each phrase, classify its relation to the image:

\begin{itemize}[leftmargin=*, itemsep=0.2mm, topsep=0.2mm]
  \item \textbf{Positive}: clearly represented in the image (e.g., anatomy, findings).
  \item \textbf{Neutral}: structural/general/meta/adjectives or not clinically informative (e.g., ``findings'', ``normal'', ``changes'').
  \item \textbf{Negative}: opposite of what is shown (e.g., concept implies pathology but image is normal).
  \item \textbf{Uncertain}: unclear presence (e.g., ``possible'', ``could be'').
\end{itemize}

\vspace{0.5mm}
\textbf{Output constraint:} Return only \textbf{Positive}, \textbf{Negative}, or \textbf{Uncertain}. Exclude \textbf{Neutral} entirely.

\vspace{0.8mm}
\textbf{Output format:}\\
\ttfamily
***start***\\
(Concept ID only (starts with C), Relation)\\
***end***\par
\normalfont

\vspace{0.8mm}
\textbf{Ensure that:}
\begin{itemize}[leftmargin=*, itemsep=0.2mm, topsep=0.2mm]
  \item Neutral concepts are excluded entirely.
  \item ``findings'' and other general/structural terms are treated as Neutral and omitted.
  \item Remaining concepts are checked against the report context and image relevance.
\end{itemize}

\end{tcolorbox}
\caption{Prompt for concept disambiguation and relation extraction.}
\label{fig:prompt}
\end{figure*}

\begin{figure*}[t]
\centering\small
\begin{tcolorbox}[
  title=Entity \& Relation Extraction Comparison Across LLMs,
  label=box:entityExtraction,
  colback=white, colframe=blue!50!black, fonttitle=\bfseries,
  width=\textwidth,
  boxsep=1mm, left=1mm,right=1mm,top=1mm,bottom=1mm
]
% \begin{tcolorbox}[float, floatplacement=h, title=Entity \& Relation Extraction Comparison Across LLMs (\textbf{E.4}\label{box:d4}), 
%                   label=box:entityExtraction, colback=white, colframe=blue!50!black, fonttitle=\bfseries]

\textbf{Radiological Report:}\\
\textit{As compared to the previous radiograph, there is a further increase in extent of the opacities in the right lung. The left lung is constant. Changed nasogastric tube. Moderate cardiomegaly with extensive retrocardiac atelectasis.}\\[2mm]
\textbf{Results:}\\
\noindent\textbf{Gemini-2.5}
\begin{quote}
\small\texttt{C0032285:Pneumonia, Uncertain\\
C0225706:Right Lung, Positive\\
C0029053:opacities, Positive\\
C0225730:Left Lung, Positive\\
C0018800:Cardiomegaly, Positive\\
C0004144:Atelectasis, Positive\\}
\end{quote}

\noindent\textbf{LLaMA-3.1}
\begin{quote}
\small\texttt{C0032285:PNEUMONIA (Pneumonia), Positive\\
C0264716:Chronic heart failure, Positive\\
C0476273:Distress, Respiratory (Respiratory distress), Positive\\}
\end{quote}

\noindent\textbf{GPT-4o}
\begin{quote}
\small\texttt{C0029053:opacities (Decreased translucency), Positive\\
C0225706:Right Lung (Right lung), Positive\\
C0085678:Nasogastric Tube (Nasogastric tube), Positive\\
C0018800:CARDIOMEGALY (Cardiomegaly), Positive\\
C0004144:ATELECTASIS (Atelectasis), Positive\\
}
\end{quote}

% \end{tcolorbox}
\end{tcolorbox}
\caption{Entity \& relation extraction comparison across LLMs.}
\label{fig:llm_comp}
\end{figure*}

\section{Deployment and Updating}

{\KG} supports three categories of updates:
\begin{enumerate}
    \item \textbf{Foundational knowledge updates} (e.g., incorporating new UMLS releases).
    \item \textbf{Imaging dataset updates} (e.g., newly added MIMIC--CXR studies or revised radiology reports).
    \item \textbf{Multimodal extensions} (e.g., integration of CT, MRI, ultrasound, or EHR-derived features).
\end{enumerate}

Because the construction pipeline is highly efficient, typically requiring only a few hours with API-based processing, the entire workflow can be re-executed whenever new data or modalities become available, thereby enabling continuous and real-time maintenance of {\KG}.

% \section{Author Statement}
% As authors of this dataset and article, we take full responsibility in the event of any violation of rights or licenses. We have included a disclaimer in the repository inviting original dataset creators to open issues regarding any license-related concerns.

% \section{Limitations}\label{app:limitations}
% Despite its contributions, {\KG} has several limitations. Due to data privacy policies and access restrictions of the MIMIC-CXR dataset (PhysioNet Credentialed Health Data License 1.5.0), sensitive medical images in {\KG} cannot be released directly along with other components. Instead, we provide the index of the images. Users are required to independently obtain the MIMIC-CXR dataset and follow our reconstruction instructions~\footnote{\url{https://anonymous.4open.science/r/MedMKG-525F/README.md}}, which may present a barrier to accessibility.

% In addition, privacy constraints prevent us from conducting detailed analyses of specific radiological images or reporting case-level findings. While aggregate analyses and quantitative evaluations are feasible, visual inspection or discussion of individual examples is restricted to comply with ethical and legal requirements.

\section{Details of Knowledge Graph Construction}
\label{app:construction}

\subsection{Pre-processing of MIMIC-CXR}
\label{app:preprocessing}
To ensure the quality of the constructed multimodal knowledge graph, we perform targeted pre-processing on the raw data in the MIMIC-CXR database. Each radiological report may correspond to images in different views, including anteroposterior, posteroanterior, lateral, etc. Involving multiple images with the same set of concepts could result in significant redundant edges within the knowledge graph. Therefore, we only maintain images in the anteroposterior view for graph conciseness; similarly, radiological reports usually contain abundant information such as diagnostic history that is not directly relevant to the content of the corresponding radiological image, therefore, extracting concepts from these similar reports can also result in redundancy. 

To mitigate this problem, we only preserve sections of Impression and Findings, two major sections that contain the most informative content, and stick to existing works in clinical report analysis~\cite{luo2024corelation}.
We perform semantic filtering using DBSCAN~\cite{ester1996density} and MedCSPCLIP~\cite{wang2024unity}. To be specific, we encode all the radiological reports with the text encoder of MedCSPCLIP, then perform clustering on the reports based on their semantics. Based on the clustering results, we select the ones near the centroid of each cluster as representative of a group of similar radiological reports.  

These approaches function together, ensuring that our pipeline referred to in Section~\ref{sec:construction} receives high-quality data for processing, producing the multimodal knowledge graph with sufficient information, negligible noise, and minimal redundancy. 

\subsection{Filtering per Semantic Type of Medical Concepts}
\label{app:semantic_type}
In order to eliminate concepts that are overly abstract or lack practical value, we filter concepts based on their semantic types. The following semantic types are not preferred and thus filtered, while all other semantic types in the UMLS vocabulary~\footnote{\url{https://www.nlm.nih.gov/research/umls/META3_current_semantic_types.html}} are allowed. Here are the filtered types:
\textit{Occupation or Discipline, Intellectual Product, Age Group, Biomedical Occupation or Discipline, Classification, Patient or Disabled Group, Organization, Regulation or Law, Geographic Area, Health Care Related Organization, Language, Conceptual Entity, Professional Society, Group Attribute, Idea or Concept, Self-help or Relief Organization, Group, Temporal Concept, Professional or Occupational Group, Qualitative Concept, Quantitative Concept, Population Group, Functional Concept, Body System, and Family Group}.                

% Figure~\ref{fig}
% \begin{table*}[t]
%     \centering
%     \caption{Filtered Semantic Types. The semantic types listed below are disallowed; all others are considered allowable.}
%     \label{tab:semantic_type}
%     % \resizebox{\columnwidth}{!}
%     {%
%         \begin{tabular}{lll}
%             \toprule
%             % \multicolumn{3}{c}{\textbf{Filtered Semantic Type}} \\
%             % \midrule
%             Occupation or Discipline            & Intellectual Product        & Age Group                \\
%             Biomedical Occupation or Discipline & Classification              & Patient or Disabled Group \\
%             Organization                        & Regulation or Law           & Geographic Area         \\
%             Health Care Related Organization    & Language                    & Conceptual Entity       \\
%             Professional Society                & Group Attribute             & Idea or Concept         \\
%             Self-help or Relief Organization    & Group                      & Temporal Concept        \\
%             Professional or Occupational Group  & Qualitative Concept         & Quantitative Concept    \\
%             Population Group                    & Functional Concept          & Body System            \\
%             Family Group                        &                            &                        \\
%             \bottomrule
%         \end{tabular}%
%     }
% \end{table*}

\subsection{Prompt for Concept Disambiguation and Relation Extraction}
\label{app:prompt}

To leverage the LLM's contextual understanding for effective concept disambiguation and relation extraction, we designed an instructive prompt that guides the model through these tasks. The prompt is presented in Figure~\ref{fig:prompt}.

\subsection{Selection of LLM}
\label{app:llm}

In this study, GPT-4o~\cite{openai2023gpt} is selected for disambiguation, as prior research has demonstrated its superior performance in biomedical comprehension
% ~\cite{silberg2024unitox, datasettowards}
and its effectiveness in resolving medical terminology ambiguity
% ~\cite{kugic2024disambiguation}
, compared with other LLMs. To further substantiate this choice, we conduct a case study to evaluate the suitability of GPT-4o for curating~\KG, as shown in Figure~\ref{fig:llm_comp}.

The analysis reveals that GPT-4o outperforms other advanced LLM backbones, i.e., Gemini-2.5~\cite{comanici2025gemini} and LLaMA-3.1-8B-Instruct~\cite{grattafiori2024llama}, by extracting more accurate concepts of interest while generating fewer hallucinations. These findings reinforce our decision to adopt GPT-4o for concept and relation extraction.

\begin{algorithm}[h]
\caption{Neighbor-Aware Filtering Algorithm}
\label{alg:cfiif}\small
\begin{algorithmic}[1]
\State \textbf{Input:} 
\begin{itemize}
    \item A set of images $\mathcal{M} = \{m_1, m_2, \dots, m_N\}$.
    \item For each image $m_i$, its associated triplets $T_i = \{(m_i, r_{ij}, c_{ij})\}$.
    \item The set of filtered clinical concepts $\mathcal{C}$.
\end{itemize}
\State \textbf{Output:} Selected image set $\mathcal{M}^*$.
\State $\mathcal{M}^* \gets \emptyset$ and $\mathcal{C}^* \gets \emptyset$.
\For{each image $m_i \in \mathcal{M}$}
    \State Compute $\displaystyle \text{Score}(m_i) \gets \sum_{(r, c) \in T_i} \log \frac{M}{|N_{(r, c)}|}$.
\EndFor
\State Sort $\mathcal{M}$ in descending order by $\text{Score}(m_i)$.
\For{each image $m_i$ in sorted order}
    \If{$\mathcal{C}^* \neq \mathcal{C}$}
        \State $\mathcal{M}^* \gets \mathcal{M}^* \cup \{m_i\}$.
        \State $\mathcal{C}^* \gets \mathcal{C}^* \cup \{ c \mid \exists\, r \text{ such that } (r, c) \in T_i \}$.
    \Else
        \State \textbf{break}
    \EndIf
\EndFor
\State \textbf{return} $\mathcal{M}^*$.
\end{algorithmic}
\end{algorithm}

\subsection{NaF Algorithm}
\label{app:NAF}
We propose the Neighbor-Aware Filtering (NaF) algorithm for effective image filtering to boost the conciseness of {\KG}. More details are presented in Algorithm~\ref{alg:cfiif}.

\subsection{Illustration of {\KG}}
\label{app:overview}

\begin{figure*}[t] 
\centering 
\begin{subfigure}[t]{0.48\textwidth} 
\centering 
\includegraphics[width=\textwidth]{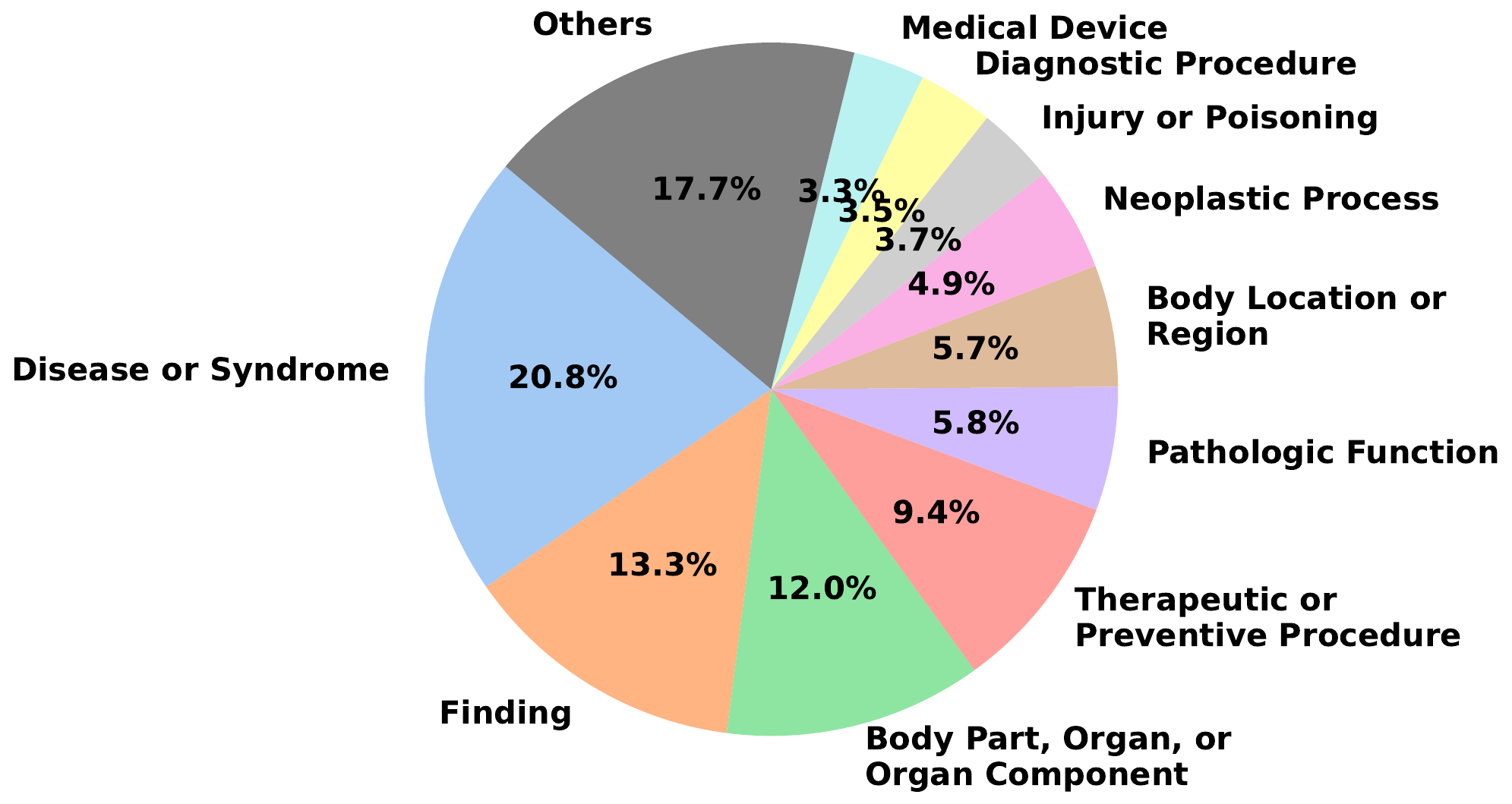} \caption{Distribution of Head Concepts per Semantic Types} \label{fig:head_entities} 
\end{subfigure} \hfill 
\begin{subfigure}[t]{0.48\textwidth} 
\centering \includegraphics[width=0.9\textwidth]{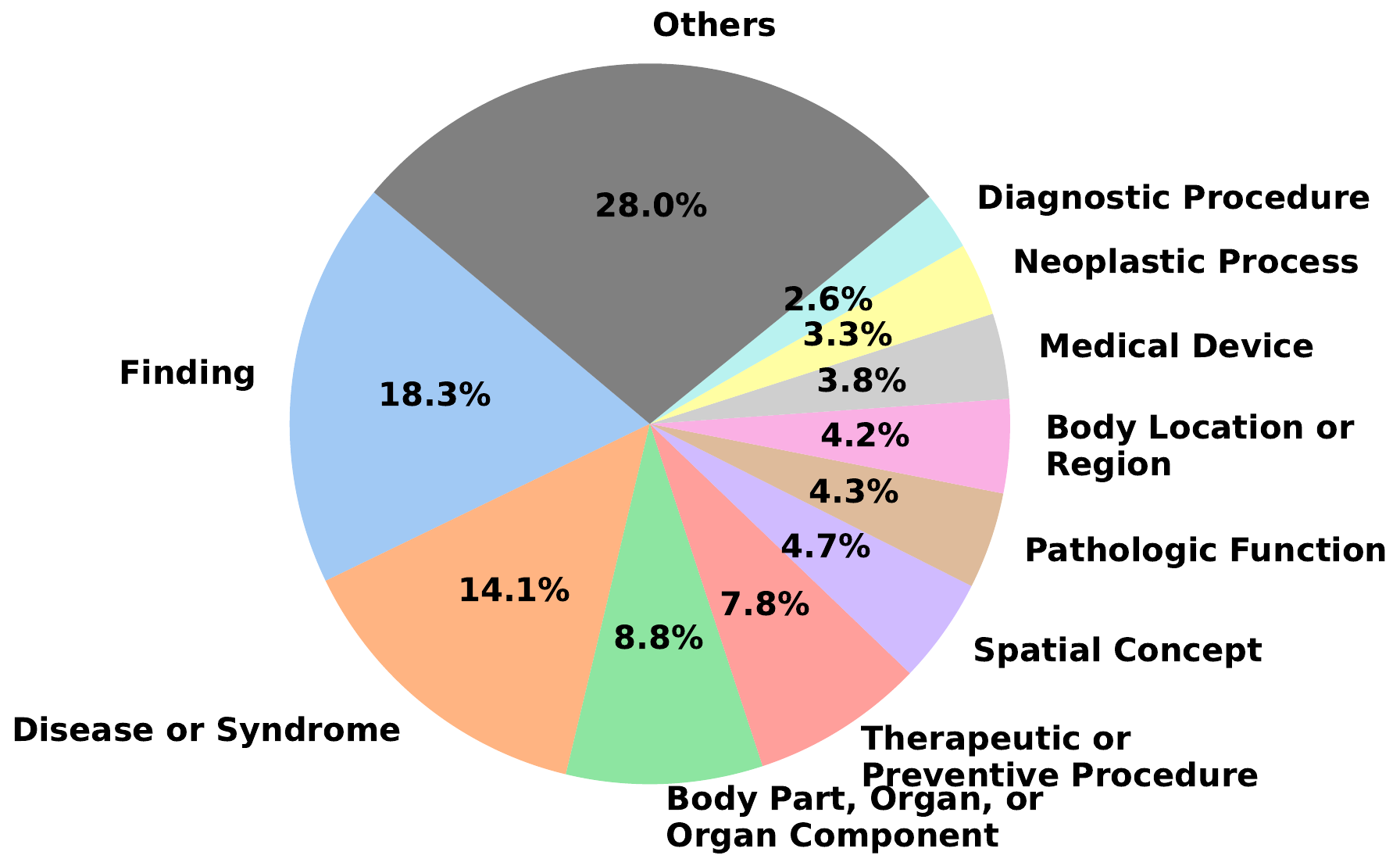} 
\caption{Distribution of Tail Concepts per Semantic Types} \label{fig:tail_entities} 
\end{subfigure} 
\caption{Distribution of entities involved in {\KG}. The top 10 semantic types are shown individually, and rare types are grouped as ``Others.''} 
% \vspace{-0.2in}
\label{fig:semantic_type} 
\end{figure*}

Figure~\ref{fig:semantic_type} shows the distribution of semantic types between the clinical concepts involved, indicating a broad and balanced coverage of the areas of clinical knowledge.
Figure~\ref{fig:demo} shows a subgraph of {\KG}, provided to facilitate a better understanding of its structure and content. As shown in Figure~\ref{fig:demo}, the medical multimodal knowledge graph integrates both intra- and cross-modal edges, offering rich multimodal medical knowledge that can potentially support a wide range of applications.

\begin{figure}[h]
    \centering
    \includegraphics[width=0.85\linewidth]{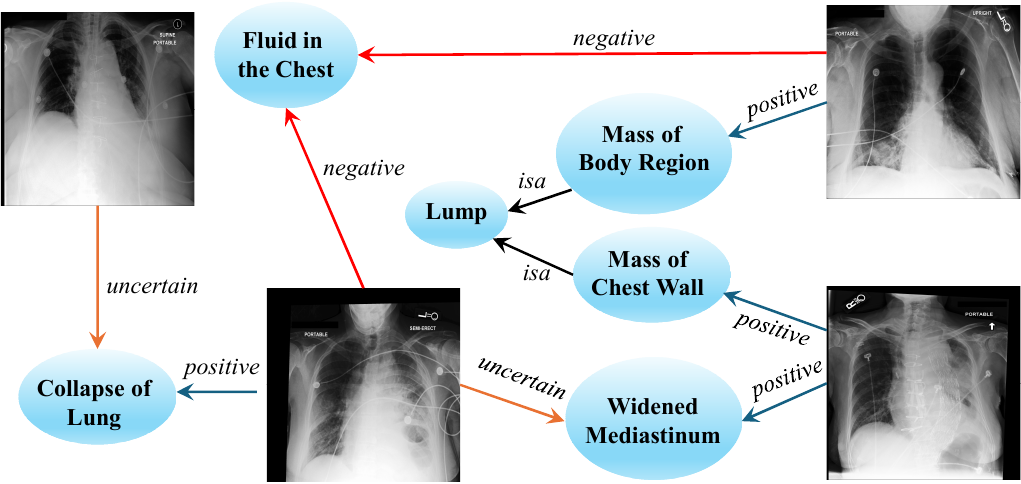}
    \caption{An illustration of {\KG}.}
    \label{fig:demo}
\end{figure}

% \end{minipage}
% \hfill
% \begin{minipage}[t]{0.48\textwidth}
% \centering
% \caption{Human assessment results.}
% \vspace{-0.1in}
% \label{fig:human_assessment}
% \includegraphics[width=0.85\textwidth]{fig/evaluation_boxplot.pdf}
% % \includegraphics[width=0.9\linewidth, height=4cm]{fig/evaluation_boxplot.eps}
% % \end{minipage}
% \vspace{-15pt}
% \end{figure*}

\section{Details of Human Assessment}
\label{app:human_asssessment}

% \subsection{Qualification of Evaluator}

% \xiaochen{TBD.}

\subsection{Assessment Criteria}

We conducted a human evaluation to assess the quality of {\KG}. Three key metrics were used:
\begin{itemize}[leftmargin=*]
\item \textbf{Concept Coverage} measures how comprehensively the extracted concepts capture the clinically meaningful findings present in the image.
\item \textbf{Relation Correctness} assesses whether the relationships between images and extracted concepts are accurately modeled, correctly identified with positive, negative, or uncertain associations.
\item \textbf{Image Diversity} evaluates whether the set of images associated with each concept reflects a diverse range of clinical scenarios, rather than highly homogeneous ones.
\end{itemize}

These metrics were selected to capture complementary aspects of performance: \textit{Concept Coverage} ensures clinical relevance and completeness; \textit{Relation Correctness} ensures accurate representation of image-concept associations; and \textit{Image Diversity:} ensures the robustness and generalizability of concept representations. Together, they provide a holistic evaluation of both precision and breadth of {\KG}.

\subsection{Assessment Procedure}

For the metrics of concept coverage and relation correctness, we randomly sample 30 images in {\KG}, choose all their concept neighbors, and the relation connecting them for assessment. For image diversity, we randomly choose 30 concepts in {\KG} and provide all the images positively linked with them to the evaluator. The evaluator performs the assessment along with detailed guidance. 

The guidebook is available in the repository.

\section{License and Access}
\label{app:license}
The MedMKG graph structure, annotations, and metadata are released under the Creative Commons Attribution 4.0 International (CC BY 4.0) license. We do not redistribute any MIMIC-CXR image files as part of MedMKG. To access the underlying images, users must obtain them separately from PhysioNet through the required credentialing process and comply with the corresponding PhysioNet Credentialed Health Data License and Data Use Agreement.

% The guidebook is available for check~\footnote{\url{https://docs.google.com/document/d/1Z--FL3-eEN4JtYosiMd07Xa_yiudQOPAuAE2wx0PNg4/edit?usp=sharing}}.

\section{Details of Link Prediction}
\label{app:link_prediction}

\subsection{Implementation}
All models are optimized using the AdamW optimizer~\cite{loshchilov2017fixing} with a batch size of 2,048 and a learning rate of 0.001. The training is run for a maximum of 500 epochs with an early stopping mechanism (patience set to 5 epochs) to prevent overfitting. Data are split into training, validation, and test sets with an 8:1:1 ratio. We run each experiment with 5 random seeds and report the mean in Table~\ref{tab:kg_baselines}.

\subsection{Link Prediction Baselines}
\label{app:link}
We benchmark {\KG} with the following baseline models in the task of link prediction:
\begin{itemize}[leftmargin=*]
    \item \textbf{AttH}~\cite{chami2020low} is a hyperbolic knowledge graph embedding model designed to capture hierarchical structures by leveraging the Lorentz model.
    % \item \textbf{RGCN}~\cite{schlichtkrull2018modeling} is a relational graph convolutional network that extends traditional GCNs to handle multi-relational knowledge graphs by learning distinct transformations per relation type.
    \item \textbf{DistMult}~\cite{yang2014embedding} is a bilinear factorization model for knowledge graphs that represents relations as diagonal matrices, enabling efficient computation.
    \item \textbf{TransR}~\cite{lin2015learning} extends TransE by introducing separate relation-specific entity spaces, allowing better modeling of diverse relationships.
    \item \textbf{HypER}~\cite{balavzevic2019hypernetwork} applies hypernetworks to generate relation-dependent transformation matrices for entity embeddings, improving flexibility.
    \item \textbf{SimplE}~\cite{kazemi2018simple} is an extension of Canonical Polyadic (CP) decomposition that enables each entity representation to be used in two different ways.
    \item \textbf{TuckER}~\cite{balavzevic2019tucker} is based on Tucker decomposition and factorizes the knowledge graph tensor into entity and relation embeddings with a core interaction tensor.
    \item \textbf{MurP}~\cite{balazevic2019multi} embeds knowledge graphs in the Poincaré ball model, enabling effective representation of hierarchical data.
    \item \textbf{MurE}~\cite{balazevic2019multi} embeds knowledge graphs in Euclidean space using multiple relational constraints to improve predictive performance.
    \item \textbf{NTN}~\cite{socher2013reasoning} introduces a neural tensor network for knowledge graph embedding, modeling entity interactions through a bilinear tensor layer.
    \item \textbf{TransD}~\cite{ji2015knowledge} extends TransE and TransH by introducing entity- and relation-specific projection matrices for dynamic embedding transformation.
    \item \textbf{TransE}~\cite{bordes2013translating} models relationships as translations in the embedding space, assuming that the sum of the head and relation embeddings approximates the tail embedding.
    \item \textbf{RESCAL}~\cite{nickel2011three} models multi-relational data using a bilinear tensor factorization approach that captures pairwise interactions.
    \item \textbf{RotatE}~\cite{sun2019rotate} represents relations as rotations in a complex vector space, capturing symmetric and antisymmetric relations effectively.
    \item \textbf{TransH}~\cite{wang2014knowledge} introduces relation-specific hyperplanes to improve the representation of diverse relational properties.
    \item \textbf{ConvE}~\cite{dettmers2018convolutional} applies 2D convolutional neural networks to entity embeddings, capturing complex interactions between entities and relations.
    \item \textbf{ComplEx}~\cite{trouillon2016complex} extends DistMult by using complex-valued embeddings, enabling the representation of asymmetric relations.
    \item \textbf{ConvR}~\cite{jiang2019adaptive} applies relation-specific convolutional filters to entity embeddings, enhancing the modeling of complex interactions.
\end{itemize}

\subsection{Evaluation Metrics}\label{app:link_prediction_metrics}

For the link prediction tasks, we use Mean Rank (MR) and Hits@K to evaluate the baselines. Lower MR and higher Hits@K indicate better performance.

% Let $\mathcal{T}$ denote the set of test triples and, for each test case $i$, let $r_i$ be the rank of the ground-truth entity among all candidate entities (with a lower rank indicating better performance). The metrics are defined as follows:

% \paragraph{Mean Rank (MR)}
% The Mean Rank is the average rank of the ground-truth entities over all test cases:
% \begin{equation}
%     \text{MR} = \frac{1}{|\mathcal{T}|} \sum_{i=1}^{|\mathcal{T}|} r_i.
% \end{equation}

% \paragraph{Mean Reciprocal Rank (MRR)}
% The Mean Reciprocal Rank is the average of the reciprocal ranks for the ground-truth entities:
% \begin{equation}
%     \text{MRR} = \frac{1}{|\mathcal{T}|} \sum_{i=1}^{|\mathcal{T}|} \frac{1}{r_i}.
% \end{equation}

% \paragraph{Hits@K}
% Hits@K measures the proportion of test cases for which the ground-truth entity is ranked within the top $K$ predictions:
% \begin{equation}
%     \text{Hits@}K = \frac{1}{|\mathcal{T}|} \sum_{i=1}^{|\mathcal{T}|} \mathbb{I}(r_i \leq K),
% \end{equation}
% where $\mathbb{I}(\cdot)$ is the indicator function that returns 1 if the condition is true and 0 otherwise.

\begin{table*}[t]
\centering
\caption{Results (\%) on knowledge-augmented visual question answering with PubMedCLIP as the backbone.}
\vspace{-0.1in}
\label{tab:vqa_pubmedclip}
\resizebox{0.73\textwidth}{!}
{%
\begin{tabular}{l|cccc|cccc|cccc}
\toprule
\multirow{2}{*}{\textbf{Methods}} & \multicolumn{4}{c|}{\textbf{VQA-RAD}} & \multicolumn{4}{c|}{\textbf{SLAKE}} & \multicolumn{4}{c}{\textbf{PathVQA}} \\
\cmidrule(lr){2-13}
 & Acc$\uparrow$ & Prec$\uparrow$ & Rec$\uparrow$ & F1$\uparrow$ & Acc$\uparrow$ & Prec$\uparrow$ & Rec$\uparrow$ & F1$\uparrow$ & Acc$\uparrow$ & Prec$\uparrow$ & Rec$\uparrow$ & F1$\uparrow$ \\
\midrule
\textbf{MedCSPCLIP} & 68.13 & 61.59 & \textbf{85.59} & 71.63 & 66.20 & 83.95 & 38.86 & 53.12 & 77.72 & 73.37 & \textbf{92.24} & 81.73 \\
$+$ KRISP & \textbf{80.08} & \textbf{84.00} & 71.19 & \textbf{77.06} & 70.70 & \textbf{91.76} & 44.57 & 60.00 & 83.19 & \textbf{94.71} & 72.96 & 82.43 \\
$+$ MKBN & 69.72 & 65.44 & 75.42 & 70.08 & 67.32 & 75.21 & 50.29 & 60.27 & 85.37 & 86.17 & 86.84 & 86.51 \\
$+$ K-PathVQA & 67.73 & 75.34 & 46.61 & 57.59 & 71.55 & 74.03 & 65.14 & 69.30 & 85.31 & 89.35 & 82.65 & 85.87 \\
$+$ EKGRL & 76.10 & 73.39 & 77.12 & 75.21 & 69.30 & 78.95 & 51.43 & 62.28 & 84.92 & 92.75 & 78.19 & 84.85 \\
$+$ MR-MKG & 78.49 & 77.59 & 76.27 & 76.92 & \textbf{83.94} & 83.15 & \textbf{84.57} & \textbf{83.85} & \textbf{86.53} & 89.74 & 84.75 & \textbf{87.17} \\
\midrule
\textbf{PubMedCLIP} & 66.14 & 64.35 & 62.71 & 63.52 & 63.94 & 59.59 & \textbf{83.43} & 69.52 & 81.26 & 86.65 & 77.20 & 81.65 \\
$+$ KRISP & 76.10 & 76.85 & 70.34 & 73.45 & 75.77 & \textbf{79.47} & 68.57 & 73.62 & 84.41 & 88.13 & 82.21 & 85.07 \\
$+$ MKBN & 67.33 & 67.31 & 59.32 & 63.06 & 70.70 & 75.18 & 60.57 & 67.09 & 84.56 & 90.15 & 80.18 & 84.87 \\
$+$ K-PathVQA & 72.51 & 76.34 & 60.17 & 67.30 & 68.17 & 67.03 & 69.71 & 68.35 & 83.76 & 87.62 & 81.44 & 84.42 \\
$+$ EKGRL & 76.49 & 75.21 & 74.58 & 74.89 & 75.49 & 73.40 & 78.86 & 76.03 & \textbf{84.59} & \textbf{90.41} & 79.96 & 84.86 \\
$+$ MR-MKG & \textbf{78.88} & \textbf{76.86} & \textbf{78.81} & \textbf{77.82} & \textbf{77.75} & 78.57 & 75.43 & \textbf{76.97} & 84.18 & 86.07 & \textbf{84.36} & \textbf{85.21} \\
\bottomrule
\end{tabular}%
}
\vspace{-0.1in}
\end{table*}

\section{Backbone Models in Knowledge-augmented Tasks}
\label{app:backbone}
The following advanced visual language models are adapted as the standard backbone for knowledge-augmented methods:
\begin{itemize}[leftmargin=*]
    \item \textbf{CLIP}~\cite{radford2021learning} is a vision-language model trained on large-scale internet data using contrastive learning. It aligns images and text embeddings in a shared latent space, enabling zero-shot image classification and retrieval. The model is under the MIT License.
    
    \item \textbf{PubmedCLIP}~\cite{eslami2023pubmedclip} is a domain-specific adaptation of CLIP trained on PubMed articles and biomedical images. It enhances the alignment of biomedical images with textual descriptions, improving zero-shot performance in medical imaging tasks. The model is under the MIT License.
    
    \item \textbf{BioMedCLIP}~\cite{zhang2023biomedclip} is a biomedical contrastive pretraining model trained on a large-scale corpus of biomedical images and text. It is designed to improve multimodal understanding in healthcare applications, particularly for retrieval and classification tasks. The model is under the MIT License.
    
    \item \textbf{MedCSPCLIP}~\cite{wang2024unity} is a medical-specific adaptation of CLIP that incorporates the MedCSP framework for contrastive scalable pretraining. It learns generalizable medical image representations, enabling improved zero-shot performance and transfer learning in clinical applications. The model is under the MIT License.
\end{itemize}

\section{Details of Knowledge-augmented Image-text Retrieval}
\label{app:kaitr}

% \subsection{Datasets}
% \label{app:kaitr_datasets}
% We utilize two prevalent datasets for benchmarking, i.e., OpenI and MIMIC-CXR. Statistics of these datasets are available at Table~\ref{}.

\subsection{Datasets \& Backbone Models}
We leverage two representative datasets for the medical text-image retrieval task, i.e., OpenI~\cite{demner2016preparing} and MIMIC-CXR~\cite{johnson2019mimic}, following previous work~\cite{wang2024unity}. To prevent any potential data leakage regarding MIMIC-CXR, we only select text-image pairs that were not used during the curation of {\KG}, and we randomly sample a fixed set of 10,000 pairs from these remaining examples. 
% Detailed dataset statistics are provided in Appendix~\ref{sec:appendix}. 
Since no predefined splits exist, both datasets are divided into training, validation, and test sets with an 8:1:1 ratio.
To comprehensively assess the impact of knowledge augmentation, we employ four open-sourced vision–language models as backbones: CLIP~\cite{radford2021learning}, PubMedCLIP~\cite{eslami2023pubmedclip}, BioMedCLIP~\cite{zhang2023biomedclip}, and MedCSPCLIP~\cite{wang2024unity}, as detailed in Appendix~\ref{app:backbone}.

\subsection{Implementation}
\label{app:kaitr_impl}
We comprehensively evaluate retrieval performance using standard metrics, i.e., precision@$K$ and recall@$K$, with $K$ set to 10, 20, and 100. Detailed metric descriptions can be found in Appendix~\ref{app:kaitr_eval}.
All models are optimized using the AdamW optimizer~\cite{loshchilov2017decoupled}. The hidden state dimension is uniformly set to 512, and the learning rate is configured to 0.0001. Training is conducted for a maximum of 30 epochs with an early-stopping patience of 3 epochs.

\subsection{Baselines}
\label{app:kaitr_baselines}
In the task of knowledge-augmented image-text retrieval, we benchmark with the following baseline models:
\begin{itemize}[leftmargin=*]
    \item \textbf{KnowledgeCLIP}~\cite{pan2022contrastive}: This model extends CLIP by integrating external knowledge graphs. By adding knowledge-based objectives during pre-training, it leverages structured relational data to improve semantic alignment between images and text.
    \item \textbf{FashionKLIP}~\cite{wang-etal-2023-fashionklip}: Designed for the fashion domain, FashionKLIP automatically constructs a multimodal conceptual knowledge graph from large-scale fashion data. By injecting domain-specific knowledge into the pre-training process, it learns fine-grained representations that enhance image-text alignment and retrieval performance.
\end{itemize}

\subsection{Evaluation Metrics}
\label{app:kaitr_eval}
For this task, we use Precision$@k$ and Recall$@k$ as evaluation metrics. A higher Precision$@k$ indicates that a larger proportion of the retrieved items are relevant, whereas a higher Recall$@k$ suggests that a greater proportion of all relevant items have been retrieved. These metrics together provide a comprehensive evaluation of the retrieval performance.

% Let $\mathcal{Q}$ denote the set of queries. For each query $q \in \mathcal{Q}$, let $R(q)$ be the set of relevant items, and let $\hat{R}_k(q)$ be the set of top-$k$ items retrieved by the model. Then, the metrics are defined as follows:

% \paragraph{Precision$@k$}
% Precision$@k$ is the fraction of the top-$k$ retrieved items that are relevant. Formally, it is given by:
% \begin{equation}
%     \text{Precision@}k = \frac{1}{|\mathcal{Q}|} \sum_{q \in \mathcal{Q}} \frac{|\hat{R}_k(q) \cap R(q)|}{k}.
% \end{equation}

% \paragraph{Recall$@k$}
% Recall$@k$ is the fraction of the relevant items that are retrieved in the top-$k$ results. It is defined as:
% \begin{equation}
%     \text{Recall@}k = \frac{1}{|\mathcal{Q}|} \sum_{q \in \mathcal{Q}} \frac{|\hat{R}_k(q) \cap R(q)|}{|R(q)|}.
% \end{equation}

\section{Details of Knowledge-augmented Visual Question Answering}
\label{app:kavqa}

% \textbf{Datasets and Backbone Models.}
% To benchmark current knowledge-augmented visual question answering methods with our proposed {\KG}, we adopt three widely used medical VQA datasets, following previous work~\cite{li2024llava}. These datasets include VQA-RAD~\cite{lau2018dataset}, Slake~\cite{liu2021slake}, and Path-VQA~\cite{he2020pathvqa}. For a fair comparison, we select closed-set questions from the datasets, which can be equally tackled by methods with different sophistication. 
% We use the same set of backbone models as in Section~\ref{sec:retrieval}, namely CLIP~\cite{radford2021learning}, PubMedCLIP~\cite{eslami2023pubmedclip}, BioMedCLIP~\cite{zhang2023biomedclip}, and MedCSPCLIP~\cite{wang2024unity}. For more details, please refer to Appendix~\ref{app:backbone}.

% \textbf{Baselines, Evaluation Metrics \& Implementation.}
% We evaluate five models that integrate knowledge graphs to enhance visual question answering: KRISP~\cite{marino2021krisp}, MKBN~\cite{huang2023medical}, K-PathVQA~\cite{naseem2023k}, EKGRL~\cite{ren2023ekgrl}, and MR-MKG~\cite{lee2024multimodal}. Detailed descriptions of these approaches are provided in Appendix~\ref{app:kavqa_baselines}.
% We adopt four widely accepted metrics for the visual question answering task: Accuracy, Precision, Recall, and F1 score. More detailed metric descriptions can be found in Appendix~\ref{app:kavqa_metrics}.
% We use the same implementation configuration as described in Section~\ref{sec:retrieval}.

\subsection{Datasets \& Backbone Models}
\label{app:kavqa_datasets}
We compare the baselines on three medical visual question answering dataset, including VQA-RAD, SLAKE and PathVQA. We extract closed questions in these datasets for benchmarking. The same set of backbone models listed in Appendix~\ref{app:backbone} are applied on the datasets.

% Statistics of these datasets are available at Table~\ref{}.

\subsection{Implementations}
We use the same implementation configuration as described in Appendix~\ref{app:kaitr_impl}.

\subsection{Baselines}
\label{app:kavqa_baselines}
In the task of knowledge-augmented visual question answering, we evaluate five models that incorporate external knowledge graphs to improve visual reasoning and answer prediction:
\begin{itemize}[leftmargin=*]
    \item \textbf{KRISP}~\cite{marino2021krisp}: This model integrates structured knowledge graphs into the VQA pipeline, refining both image representations and question understanding to boost answer accuracy.
    \item \textbf{MKBN}~\cite{huang2023medical}: Originally designed for medical VQA, MKBN leverages domain-specific knowledge graphs to align visual and textual features, thus enhancing performance in specialized settings.
    \item \textbf{K-PathVQA}~\cite{naseem2023k}: By incorporating multi-hop reasoning over a knowledge graph, K-PathVQA enables the model to infer complex relationships and answer questions that require multi-step deductions.
    \item \textbf{EKGRL}~\cite{ren2023ekgrl}: This framework combines graph-based representation learning with reinforcement learning to effectively integrate external knowledge, thereby improving reasoning capabilities in visual question answering.
    \item \textbf{MR-MKG}~\cite{lee2024multimodal}: MR-MKG utilizes contrastive loss to capture diverse semantic interactions between visual content and questions, leading to enhanced cross-modal alignment and VQA performance.
\end{itemize}

\subsection{Evaluation Metrics}
\label{app:kavqa_metrics}
For the visual question answering task, we adopt four standard metrics: Accuracy, Precision, Recall, and F1 score. 

% Let $\mathcal{D}$ denote the set of VQA examples. For each example $i \in \mathcal{D}$, let $y_i$ be the ground-truth answer and $\hat{y}_i$ the predicted answer. The metrics are defined as follows:

% \paragraph{Accuracy}
% Accuracy measures the proportion of correctly answered questions:
% \begin{equation}
%     \text{Accuracy} = \frac{1}{|\mathcal{D}|} \sum_{i \in \mathcal{D}} \mathbb{I}(\hat{y}_i = y_i),
% \end{equation}
% where $\mathbb{I}(\cdot)$ is the indicator function.

% \paragraph{Precision}
% Precision is the fraction of true positive answers among all answers predicted as positive. In a binary (or thresholded) setting, it is given by:
% \begin{equation}
%     \text{Precision} = \frac{TP}{TP + FP},
% \end{equation}
% with $TP$ and $FP$ denoting the numbers of true positives and false positives, respectively.

% \paragraph{Recall}
% Recall is defined as the fraction of true positive answers among all actual positive answers:
% \begin{equation}
%     \text{Recall} = \frac{TP}{TP + FN},
% \end{equation}
% where $FN$ represents false negatives.

% \paragraph{F1 Score}
% The F1 score is the harmonic mean of Precision and Recall:
% \begin{equation}
%     \text{F1} = 2 \cdot \frac{\text{Precision} \cdot \text{Recall}}{\text{Precision} + \text{Recall}}.
% \end{equation}

% Together, these metrics provide a comprehensive evaluation of model performance on the knowledge-augmented visual question answering task.

\subsection{Additional Experiment Results}

In Table~\ref{tab:vqa_pubmedclip}, we provide additional experiment results on the knowledge-augmented visual question answering task with the backbones of MedCSPCLIP and PubMedCLIP, which are not covered in Table~\ref{tab:vqa_baselines} due to space limitations. 

Consistent with Table~\ref{tab:vqa_baselines}, despite minor instability, incorporating {\KG} yields notable performance improvements across the baselines. This trend reinforces our conclusion that the graph’s clinical specificity, image-aware relational structure, and semantic richness jointly contribute to stronger multimodal understanding.

% \section{Details of Knowledge Graph Construction}
% \label{app:kg}

\section{Ethics Considerations}

This work is guided by the principles of contributing to human well-being and avoiding harm. While {\KG} is intended to advance socially responsible and equitable research, we acknowledge potential risks such as diagnostic errors, biased decision support, or reinforcement of health disparities if models trained on it are misused. To minimize such harms, we stress the importance of expert validation, continuous monitoring of deployed systems, and safeguards against unverified clinical use. We further encourage broad, responsible accessibility of the resource, prioritizing the needs of less advantaged groups and ensuring that its use respects diversity, privacy, and safety across socio-economic contexts.

\end{document}